\documentclass{article}
\usepackage{spconf}
\usepackage{amsmath,amssymb}
\usepackage{booktabs}
\usepackage{graphicx}
\usepackage{cite}
\usepackage{url}
\usepackage{array}
\usepackage{hyperref}

\ninept
\begin{document}

\title{Attributing Preprocessing Invariance in Spectral Foundation Models}

\name{Dongjun Wei$^{1,2}$ \qquad Hongyi Wu$^{2,3}$}
\address{$^1$ESCP Business School, Paris \quad $^2$OpluxCare Tech Group Ltd. \\
$^3$The University of Hong Kong, Hong Kong}

\maketitle

\begin{abstract}
A spectral foundation model should remain useful when laboratories preprocess
spectra differently. The standard test trains a classifier under one pipeline
and evaluates under another, taking preserved accuracy as evidence of learned
invariance. However, these models normalize each input before any learned
parameter is applied. When normalization maps differently preprocessed spectra
to the same vector, the encoder receives identical inputs and the measured
invariance cannot be attributed to learning. We propose a normalization-only attribution control:
compare the encoder against its normalization before interpreting transfer as
learned invariance. On six Raman datasets the encoder does not measurably
improve transfer over its normalization. A controlled experiment confirms
invariance develops only when variation reaches the encoder past normalization.
Across three systems, no encoder improves relative retention over its
normalization. An audit of eighteen configurations across five modalities confirms
the issue is widespread: the normalization-only control should be
reported before crediting transfer to the encoder.
\end{abstract}


\section{Introduction}
\label{sec:intro}

A spectral foundation model should remain useful whichever laboratory
produced its input, since laboratories preprocess
differently~\cite{poth2022preproc}. The standard robustness test trains a
classifier under one preprocessing pipeline and evaluates under
another~\cite{rspte,dscf2025}, interpreting preserved accuracy as evidence
that the encoder learned invariance. We show this test can misattribute
invariance to the encoder when the model's own normalization is responsible.

\begin{figure}[!t]
\centering
\includegraphics[width=\columnwidth]{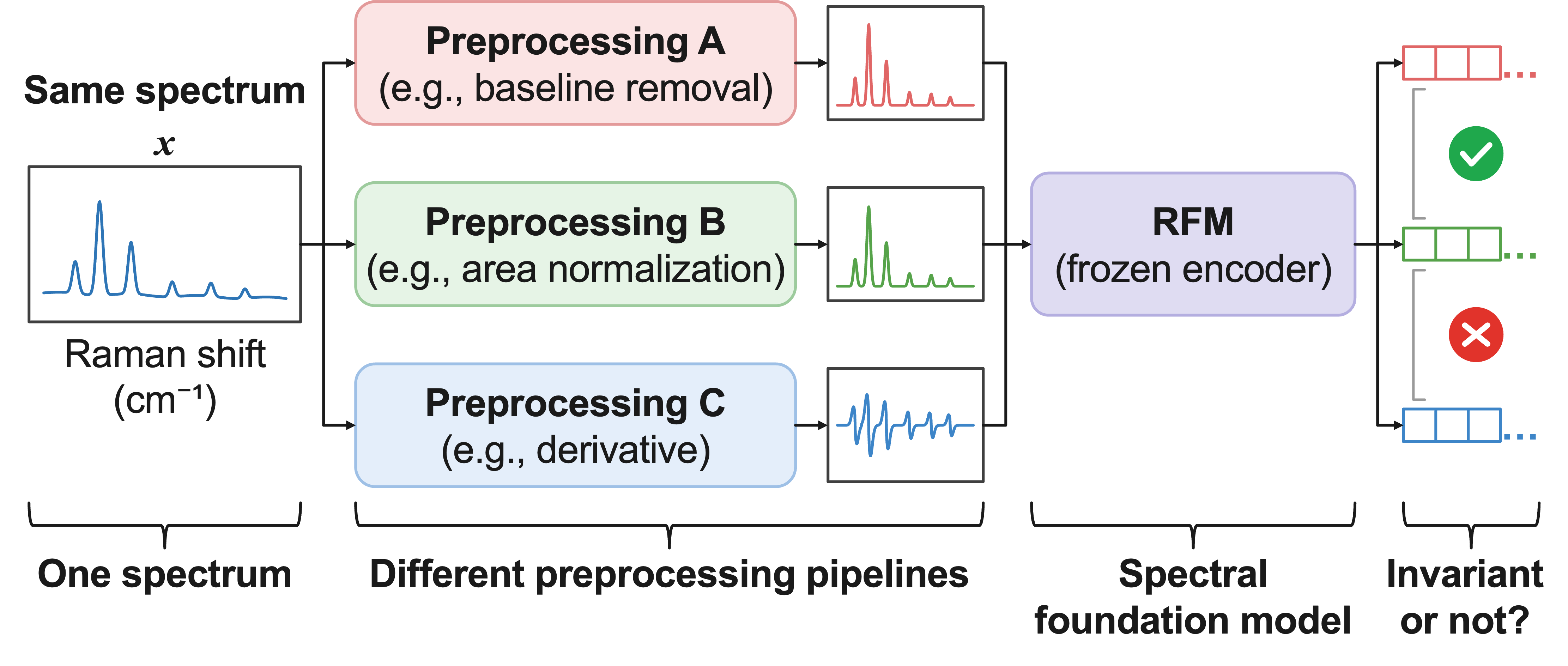}
\caption{Normalization accounts for preprocessing invariance. One spectrum
processed three ways. Baseline removal~(A) and area normalization~(B) only
shift or scale the signal, so normalization maps both to the same vector as
the original and the encoder receives identical inputs regardless of
training. Differentiation~(C) changes the signal shape and survives
normalization.}
\label{fig:intro}
\end{figure}

Most such models normalize each input before the encoder sees it.
Figure~\ref{fig:intro} illustrates: one Raman spectrum is processed three
ways. Baseline removal and area normalization only shift or scale the signal,
so normalization maps both to the same vector as the original and the encoder
receives identical inputs regardless of training. The measured invariance
belongs to the normalization, not the encoder. Only a preprocessing change
that survives normalization can test whether the encoder learned anything.

Write the model as $f(x)=g(R(x))$, where $R$ is the normalization and $g$ the
encoder. If $R(T(x))=R(x)$, then $f(T(x))=f(x)$ for every $g$: transfer
across $T$ is a property of $R$, not $g$. We propose a normalization-only
attribution control: compare the encoder against its normalization alone
before interpreting transfer as learned invariance.

We test this principle in three parts. \textbf{\textit{First}}, for a
normalization that centers and scales by per-sample statistics, the
transformations it removes are exactly the positive-affine maps
(Section~\ref{sec:removes}). \textbf{\textit{Second}}, on six Raman datasets
the encoder does not measurably outperform its normalization, and replication
on two other systems confirms no improvement in relative retention
(Section~\ref{sec:eval}). \textbf{\textit{Third}}, controlled experiments
confirm the mechanism (Section~\ref{sec:controls}): an injection-point
intervention shows that invariance develops only when variation reaches the
encoder past the normalization. Training does improve the encoder over random
initialization. Without the normalization control, cross-preprocessing
transfer can be credited to the learned encoder even when the normalization is
the actual source.

\subsection{Related work}

Per-sample normalization is long established:
SNV~\cite{barnes1989snv}, RNV~\cite{guo1999rnv}, and
MSC~\cite{geladi1985msc} are standard tools with known algebraic
links~\cite{dhanoa1994link}. The same normalization is standard in
speech~\cite{wav2vec2,hubert,wavlm}, ECG~\cite{hubertecg,ecgfm}, and
EEG~\cite{bendr} encoders. View-based
self-supervision~\cite{chen2020simclr,zbontar2021barlow} and its invariances
have been analyzed for images~\cite{ericsson2021invariance}. Lin and
Goto~\cite{lin2019zeromean} compare frame normalization and learned mechanisms
for sound-level invariance, but ask how to obtain robustness rather than
whether observed robustness was supplied by normalization. EEG foundation
models have been tested against random
initialization~\cite{zare2026eeg}, but that control requires a trained model.
We are not aware of prior work that compares a model against its own
normalization to separate learned from normalization-supplied invariance.

\section{Which Transformations the Normalization Removes}
\label{sec:removes}

\subsection{The model and its normalization}

We study a Raman foundation model (RFM) with 2.68M parameters, a transformer
pretrained on 815k spectra with masked reconstruction~\cite{he2022mae},
redundancy reduction~\cite{zbontar2021barlow}, and domain-adversarial
training~\cite{ganin2016dann}. Its first operation on every input is RNV
normalization:
\begin{equation}
R(x)=\mathrm{clip}\!\left(\frac{x-\mathrm{med}(x)\mathbf{1}}{s(x)+\epsilon},\,-8,8\right)\!,\;\epsilon{=}10^{-6}\!,
\label{eq:rnv}
\end{equation}
where $s(x)$ is half the distance between the 16th and 84th percentiles. RNV
is the deterministic part of the model before any learned parameter and so
serves as the natural comparison for isolating the encoder's contribution.
Because it produces one feature per channel at full width, it is an
attribution comparison, not a proposed deployment substitute.

\subsection{Characterization}

We write $\tilde R$ for the \emph{ideal normalization} that sets $\epsilon=0$
and removes the clipping.

\noindent\textbf{Proposition 1.}
\textit{Let $\tilde R(x)=(x-m(x)\mathbf{1})/s(x)$ be defined on inputs with
$s(x)>0$, where $m(ax+b\mathbf{1})=am(x)+b$ and $s(ax+b\mathbf{1})=as(x)$ for
$a>0$, $b\in\mathbb{R}$. Then $\tilde R(y)=\tilde R(x)$ iff
$y=ax+b\mathbf{1}$ for some $a>0$, $b\in\mathbb{R}$.}
The converse reconstructs both inputs from $z=\tilde R(x)$ as
$x=s(x)z+m(x)\mathbf{1}$ and $y=s(y)z+m(y)\mathbf{1}$, giving
$y=(s(y)/s(x))x+c\mathbf{1}$, and the coefficient $s(y)/s(x)$ is positive because
both scales are positive by assumption.
The proposition concerns the idealized normalization $\tilde R$. All claims for
the deployed operator $R$ are numerical effective-removal claims.

SNV, area normalization with positive divisor, and MSC with positive fitted
slope all apply positive-affine maps, so $\tilde R$ removes them exactly.
Derivatives, baseline removal, smoothing, and per-channel multiplicative gains
lie outside this class. RNV, $z$-scoring, and min-max normalization all satisfy
the assumptions. Throughout the paper, \emph{exactly removed} refers to
$\tilde R$, while the deployed $R$ in~(\ref{eq:rnv}) \emph{effectively
removes} these transformations: $d_R<10^{-3}$ for more than 99\% of
evaluation spectra. Verdicts are unchanged from $10^{-4}$ to
$10^{-1}$ because removed and surviving transformations are separated by five
orders of magnitude.

\subsection{Numerical test and cross-modality evidence}

The characterization can be tested numerically: normalize a sample and its
transformed version and compare. Define $d_R(x,T)=\|R(T(x))-R(x)\|_\infty$.
A transformation counts as effectively removed when $d_R<10^{-3}$ for at
least 99\% of evaluation spectra. The test depends only on the normalization
and the transformation, so it can be run before training. On 4,998 ECG time series across eight transformations and five
normalizations, verdicts match the algebraic prediction.

We ran this test on eighteen released configurations across five
modalities~\cite{rspte,dscf2025,hubertecg,ecgfm,wav2vec2,hubert,wavlm,bendr}.
Twelve normalizations remove at least part of the positive-affine variation. In
four the normalization sits between the evaluated transformation and the
learned parameters, so the standard comparison is insufficient for
attribution. Per-sample normalization appears across spectral, audio, ECG,
EEG, and PPG pipelines. Where an affine augmentation is also present, the
relevant distinction is where it is applied: one PPG model applies gain
augmentation after normalization, where it reaches the encoder, while device
gain arriving before normalization at inference is removed. The numerical
test detects this distinction before any training begins.

\section{Evaluation}
\label{sec:eval}

Section~\ref{sec:removes} identified which transformations the normalization
removes in theory. We now test whether this matters in practice: does the
encoder add anything beyond what normalization already provides?

\subsection{Setup}
\label{sec:setup}

\par\noindent\textbf{Datasets.}\enspace
We evaluate on six public Raman classification datasets (3,291 spectra total,
two to twelve classes per dataset, four from
RamanBench~\cite{ramanbench}).

\par\noindent\textbf{Preprocessing.}\enspace
Six pipelines are applied. Three are in-family, meaning the model saw them
during pretraining: raw spectra, the rolling-minimum baseline as released with
the model, and a central first difference. Three are out-of-family: asymmetric
least squares baseline~\cite{eilers2005baseline}, MSC against the source-training-fold
mean~\cite{geladi1985msc}, and Savitzky\mbox{-}Golay first
derivative~\cite{savitzky1964}.

\par\noindent\textbf{Probing protocol.}\enspace
We compare raw spectra, RNV alone, and the frozen RFM checkpoint. The probe is
feature standardization followed by logistic regression under five-fold
stratified cross-validation over three seeds. Standardization is fitted on the
training fold and applied unchanged to the target. Cell $(i,j)$ of each
transfer grid reports balanced accuracy trained under pipeline $i$, evaluated
under $j$.

\par\noindent\textbf{Metrics and inference.}\enspace
The primary endpoint is mean off-diagonal balanced accuracy (transfer),
averaged within each dataset then across six datasets. Degradation is
on-diagonal minus off-diagonal (lower is better). We additionally report
retention, off-diagonal divided by on-diagonal (higher is better), as a
scale-normalized measure of relative preprocessing robustness that factors
out within-pipeline representation quality. All intervals are paired bootstrap, 10,000 resamples, percentile
method.

\subsection{Main result}
\label{sec:main}

\begin{table}[t]
\caption{Main comparison on six Raman datasets (\%), averaged over three seeds.
$\Delta$: RFM minus RNV in points. Higher is better except degradation (lower
is better). Normalization is ahead on five of six datasets.}
\label{tab:main}
\centering
\footnotesize
\setlength{\tabcolsep}{3.5pt}
\begin{tabular}{@{}l cc @{\hskip 6pt} c c@{}}
\toprule
Metric & RNV & RFM & $\boldsymbol{\Delta}$ & 95\% CI\\
\midrule
\textbf{Transfer}  & \textbf{52.0} & 50.5 & $-1.5$ & $[-4.6,\!+\!3.3]$\\
On-diagonal         & \textbf{94.0} & 90.9 & $-3.1$ & \\
Degradation         & 42.0 & 40.4 & $-1.6$ & $[-5.4,\!+\!1.8]$\\
Retention           & 55.3 & 55.6 & $+0.2$ & $[-2.9,\!+\!4.6]$\\
\bottomrule
\end{tabular}
\end{table}

\begin{figure*}[t]
\centering
\includegraphics[width=0.95\textwidth]{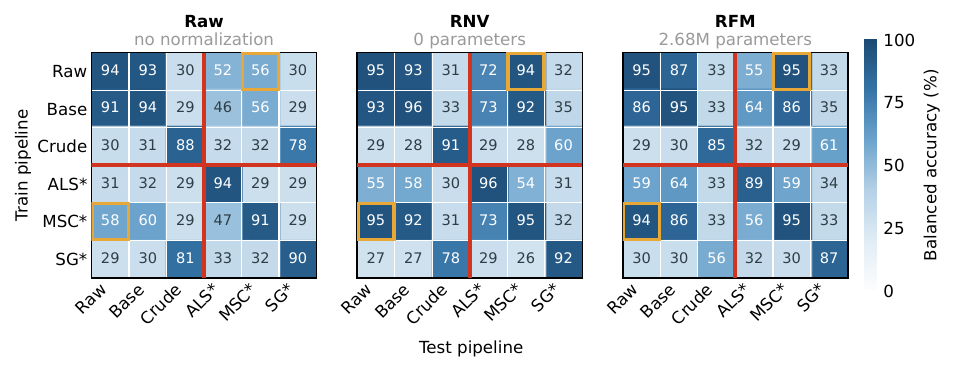}
\caption{Transfer grids. Balanced accuracy (\%) for a probe trained on pipeline
$i$ and evaluated on pipeline $j$, averaged over six datasets. Asterisks mark
out-of-family pipelines. Gold borders mark the two cells whose transfer the
ideal normalization exactly removes (raw$\leftrightarrow$MSC). Normalization
reproduces most of the transfer structure. The encoder is substantially weaker
on derivative transfer.}
\label{fig:grids}
\end{figure*}

Table~\ref{tab:main} and Figure~\ref{fig:grids} show that RFM does not
improve cross-preprocessing transfer over its own normalization.
The paired difference is $-1.5$ points $[-4.6,+3.3]$, so the interval does not
support an RFM advantage. Normalization is ahead on five of six datasets. The
conclusion is unchanged under a radial-basis probe ($-0.3$) or with
source-only penalty selection ($-1.2$).

\par
\begin{figure}[t]
\centering
\includegraphics[width=0.85\columnwidth]{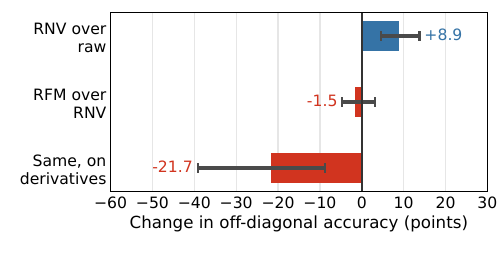}
\caption{What each stage contributes. Change in off-diagonal balanced accuracy
(points) with paired bootstrap intervals. Raw$\to$RNV: $+8.9$.
RNV$\to$RFM: $-1.5$. The model improves on raw spectra, but the improvement
is already reproduced by normalization alone.}
\label{fig:decomp}
\end{figure}

\noindent\textbf{Where the apparent advantage came from.}\enspace
Under the original evaluation protocol, RFM shows an 18-point retention
advantage over raw spectra. That protocol included cells whose transfer the
normalization removes.
Excluding the removed cells eliminates the gap ($-1.2$).
Figure~\ref{fig:decomp} separates the increments: normalization contributes
$+8.9$ points of off-diagonal accuracy, while the trained encoder changes
transfer by $-1.5$ points.

\begin{figure}[t]
\centering
\includegraphics[width=0.85\columnwidth]{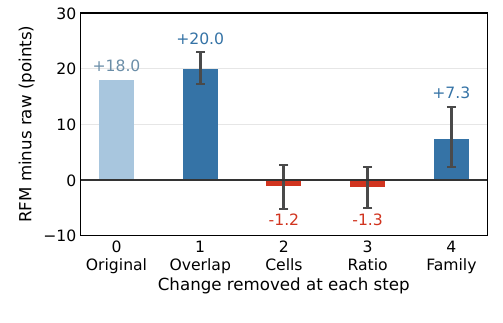}
\caption{The gap under successive protocol changes. RFM minus raw spectra
(points) at each step: original protocol, excluding shared-dataset overlap,
excluding cells the ideal normalization exactly removes, replacing the
retention ratio with absolute transfer difference, restricting to
out-of-family pipelines only. The gap disappears when the removed cells are
excluded.}
\label{fig:staircase}
\end{figure}

Figure~\ref{fig:staircase} traces how the reported advantage changes under
successive corrections. Removing dataset overlap does not reduce the gap,
which rises slightly to $+20.0$. The change occurs when cells the ideal
normalization removes are excluded: the gap falls to $-1.2$. On
out-of-family pipelines alone, an absolute gap of $+7.3$ remains over raw
spectra, but measured against the normalization rather than raw spectra, RFM
differs by only $-0.4$ points.

\par\noindent\textbf{In-family versus out-of-family.}\enspace
The six pipelines split into three in-family (seen during pretraining) and
three out-of-family (unseen). On in-family pairs, normalization and RFM are
effectively tied. On out-of-family pairs alone, RFM shows a modest absolute
advantage over raw spectra ($+7.3$ points) but no advantage over its own
normalization ($-0.4$). The distinction matters: in-family transfer may benefit from exposure to the corresponding
preprocessing during pretraining, while out-of-family
transfer tests generalization. That the normalization-only control matches
RFM on both confirms the finding is not an artifact of pipeline familiarity.

\subsection{Replication on two released systems}
\label{sec:external}

The main result could be a property of one model. We run the same comparison on
DSCF~\cite{dscf2025} (397M parameters, min-max normalization) and
HuBERT-ECG~\cite{hubertecg} (30.5M parameters, per-lead min-max
normalization, five-way superclass on PTB-XL with the patient-disjoint split).
Table~\ref{tab:external} gives the result. On relative retention, none of the
three encoders improves detectably over its normalization, at $+0.2$, $-1.8$,
and $-19.2$ points. HuBERT-ECG is $+10.8$ points better on absolute transfer
yet $-19.2$ worse on retention: the encoder lifts representation quality while
retaining a smaller fraction across a preprocessing change.
The pattern is consistent across architectures and modalities: all three use
per-sample normalization as their first operation, and in none does the
encoder detectably improve relative retention over that normalization.

\begin{table}[t]
\caption{Replication on two released systems. $\boldsymbol{\Delta}$ with 95\%
bootstrap CI in brackets. Intervals resample the six Raman datasets for
RFM and DSCF, and held-out patients for HuBERT-ECG.}
\label{tab:external}
\centering
\footnotesize
\setlength{\tabcolsep}{1.8pt}
\begin{tabular}{@{}l ccc@{}}
\toprule
Contrast & RFM & DSCF & HuBERT-ECG\\
\midrule
Norm $-$ raw       & $\!+\!8.9\,[4.6,\!13.8]$   & $\!+\!8.9\,[2.5,\!15.9]$    & $\!+\!4.4\,[3.7,\!5.2]$\\
Model $-$ norm     & $\!-\!1.5\,[-4.6,\!3.3]$    & $\!-\!5.4\,[-12.5,\!-\!1.0]$  & $\!+\!10.8\,[9.7,\!12.0]$\\
\quad $\Delta$ retention    & $\!+\!0.2\,[-2.9,\!4.6]$    & $\!-\!1.8\,[-6.8,\!2.7]$    & $\!-\!19.2\,[-23.6,\!-\!15.8]$\\
\bottomrule
\end{tabular}
\end{table}

\section{Attribution Controls and Intervention}
\label{sec:controls}

The evaluation shows the encoder does not beat normalization on transfer. We
now ask why, ruling out the dimensionality mismatch and testing the causal
mechanism directly.

To quantify how much a preprocessing change moves the representation, we
define relative feature distance as the median over spectra of
$\|h(T_2(x))-h(T_1(x))\|_2/\|h(T_1(x))\|_2$
for a representation $h$ and two pipelines $T_1,T_2$. The ratio cancels the
scale of the representation.

\par\noindent\textbf{Untrained encoder.}\enspace
If the normalization maps a transformed input and the original to the same
vector, every encoder is invariant, trained or not. On
transformations that $\tilde R$ exactly removes, relative feature distance is
near machine precision for all three: $1.9{\times}10^{-5}$ (normalization),
$5.7{\times}10^{-6}$ (untrained encoder, ten seeds), and
$1.7{\times}10^{-6}$ (trained encoder).

\par\noindent\textbf{Width-matched controls.}\enspace
The encoder outputs 320 dimensions while normalization keeps up to 2,928
channels. To test whether this dimensionality gap explains the result,
Table~\ref{tab:matched} compares same-width controls on the cells whose
transfer the normalization does not remove. A 320-dimensional random
projection of RNV matches or slightly exceeds RFM transfer ($50.6$ vs.\
$47.4$, difference $-3.2$, 95\% CI $[-5.9,+0.1]$). The interval
marginally includes zero, so RFM is not clearly worse, but the
dimensionality objection does not hold. The trained encoder outperforms its
untrained self by $+5.4$ points $[+1.9,+8.9]$, the clearest evidence that
training helps. PCA fitted on raw spectra tracks variation that the evaluated
pipelines exist to remove.

\begin{table}[t]
\caption{Controls matched to the encoder's output width (320 dim.). Balanced
accuracy (\%) on the cells whose transfer is not removed by normalization.
$\Delta$: RFM minus that row. Random projection and untrained encoder are each
averaged over ten draws.}
\label{tab:matched}
\centering
\footnotesize
\setlength{\tabcolsep}{2.5pt}
\begin{tabular}{@{}lcccc@{\hskip 4pt}cc@{}}
\toprule
Representation & Lrn. & Dim & On-d. & Off-d. & $\boldsymbol{\Delta}$ & 95\% CI\\
\midrule
RNV full       & no  & ${\le}$2928 & \textbf{94.0} & \textbf{49.0} & $-1.6$ & $[-5.0,\!+\!3.4]$\\
RNV+PCA        & no  & 320 & 83.2 & 41.5 & $+5.9$ & $[+0.9,\!+\!11.2]$\\
RNV+rand.proj. & no  & 320 & 93.3 & 50.6 & $-3.2$ & $[-5.9,\!+\!0.1]$\\
RNV+untrained  & no  & 320 & 87.8 & 42.0 & $+5.4$ & $[+1.9,\!+\!8.9]$\\
\midrule
RFM (trained)  & yes & 320 & 90.9 & 47.4 &  & \\
\bottomrule
\end{tabular}
\end{table}

\par\noindent\textbf{Injection-point intervention.}\enspace
We trained a two-layer transformer (128 dim., same RNV front end,
masked-reconstruction objective, 300 steps) under two conditions differing
only in where a random positive-affine map
$x\mapsto ax+b\mathbf{1}$, $a{\sim}\mathrm{Uniform}(0.5,1.5)$,
$b{\sim}\mathrm{Uniform}(-0.3,0.3)$, is applied: before normalization
(removed) or after it (reaches the encoder).
For evaluation, we apply the affine perturbation after normalization for both
trained models, so the measured distance tests encoder-level invariance rather
than invariance supplied by RNV. Over five paired seeds, only the condition in
which the variation reaches the encoder during training develops learned
invariance: relative feature distance falls from $0.334$ to $0.041$, versus
$0.334$ to $0.336$ in the first. Training losses decline smoothly in both, so
watching only the loss would suggest invariance was learned in both cases.

\par\noindent\textbf{Augmentation overlap.}\enspace
The pretraining augmentations perturb six physical components. Per-sample gain
and offset are fully absorbed by RNV, with relative separations of
$4.9{\times}10^{-6}$ and $2.0{\times}10^{-6}$, while fluorescence baseline,
wavenumber shift, scatter slope, and shot noise pass through at $0.03$ to
$5.4$. RNV reduces gain and offset perturbations to near numerical precision, so
these augmentations provide essentially no corresponding variation to the
encoder. This is consistent with the encoder failing to add robustness for
transformations that normalization already removes.

\par\noindent\textbf{Contraction is not transfer.}\enspace
Between the crude and the Savitzky\mbox{-}Golay derivative, the trained
encoder reduces relative feature distance threefold ($0.163$ to $0.054$),
evidence of learned feature-space contraction across the derivative pair. Yet
transfer is worst here ($-21.7$ points,
negative on all six datasets). Retention is lower for RFM ($68.5$ vs.\
$75.2$), and the loss is directional: RFM transfers comparably from crude to
Savitzky\mbox{-}Golay ($61.2$ vs.\ $59.9$) but much worse in reverse ($56.1$
vs.\ $77.8$). Representation contraction and decision-boundary preservation are distinct
properties. The derivative result is a qualification rather than a
contradiction: the encoder does learn something about the derivative pair,
but what it learns does not translate into improved probe transfer.

\par\noindent\textbf{Tokenizer effects.}\enspace
The tokenizer summarizes patches by statistics including the mean and standard
deviation. Differentiation roughly zeros the mean and rescales the spread, so
the derivative deficit may reflect this tokenizer receiving an input it was not
designed for. This does not affect the attribution argument but limits how far
the result extends beyond this architecture.

\section{Conclusion}
\label{sec:conclusion}

High cross-preprocessing transfer is not by itself evidence of learned
preprocessing invariance, because a deterministic normalization can impose that
invariance before any learned parameter is applied. Across three systems, no
encoder measurably improves relative retention over its normalization.
Recommendations: (1)~identify which transformations the normalization removes,
(2)~report the normalization-only attribution control, (3)~run the numerical
test before training, since an augmentation removed before the encoder cannot itself provide
variation from which to learn invariance to that augmentation.

We do not claim the encoder learns no invariance. The encoder outperforms its
untrained self on transfer that normalization does not remove. What we add is
an attribution principle: credit preprocessing robustness to learned
parameters only after ruling out invariance imposed by the deterministic
preprocessing. The proposed control requires no retraining, adds a single row
to existing evaluation tables, and can be applied to any model with
deterministic preprocessing. The numerical test requires only the
normalization code and a sample of inputs and can be run before training.

\par\noindent\textbf{Limitations.}\enspace
The confidence interval excludes an average RFM advantage larger than $3.3$
points but does not rule out smaller effects. Several contrasts whose
bootstrap intervals exclude zero do not do so under a paired $t$ interval.
All Raman comparisons are within-dataset, and the regime the encoder is
built for, one probe shared across instruments, remains untested because
the six datasets have disjoint label spaces. The derivative deficit may
reflect tokenizer design. Frozen probes measure transfer differently from
fine-tuning, and the absolute numbers should be read as transfer under this
protocol rather than as estimates for deployment. The audit shows the
attribution issue exists in released systems but establishes possibility, not
prevalence. Width-matched controls depend on the reduction method. Other
architectures or training objectives may yield different conclusions.

\newpage
\bibliographystyle{IEEEbib}
\bibliography{refs}
\end{document}